\documentclass{article}

\usepackage{arxiv}

\usepackage[utf8]{inputenc} 
\usepackage[T1]{fontenc}    
\usepackage{hyperref}       
\usepackage{url}            
\usepackage{booktabs}       
\usepackage{amsfonts}       
\usepackage{nicefrac}       
\usepackage{microtype}      
\usepackage{graphicx}
\usepackage{hyperref}
\usepackage[sort&compress,numbers]{natbib}
\usepackage{xcolor}
\usepackage{colortbl}

\definecolor{Gray}{gray}{0.83}
\newcolumntype{a}{>{\columncolor{Gray}}c}
\newcolumntype{b}{>{\columncolor{white}}c}

\newcommand{\absdiv}[1]{%
	\par\addvspace{.5\baselineskip}
	\noindent\textbf{#1}\quad\ignorespaces
}

\title{m2caiSeg: Semantic Segmentation of Laparoscopic Images using Convolutional Neural Networks}

\author{
  Salman Maqbool\thanks{Corresponding author - Email: smaqbool@smme.edu.pk, ORCID: 0000-0003-4120-9401} \\
  School of Mechanical and Manufacturing Engineering\\
  National University of Sciences and Technology\\
  Islamabad, Pakistan \\
   \And
 Aqsa Riaz \\
  School of Mechanical and Manufacturing Engineering\\
  National University of Sciences and Technology\\
  Islamabad, Pakistan \\
   \And
 Hasan Sajid \\
  School of Mechanical and Manufacturing Engineering\\
  National University of Sciences and Technology\\
  Islamabad, Pakistan \\
  National Center for Artificial Intelligence\\
  Islamabad, Pakistan \\
   \And
 Osman Hasan \\
  School of Electrical Engineering and Computer Science, \\
  National University of Sciences and Technology, \\
  Islamabad, Pakistan \\
}

\begin{document}
\maketitle

\begin{abstract}
\absdiv{Purpose}
Autonomous surgical procedures, in particular minimal invasive surgeries, are the next frontier for Artificial Intelligence research. However, the existing challenges include precise identification of the human anatomy and the surgical settings, and modeling the environment for training of an autonomous agent. To address the identification of human anatomy and the surgical settings, we propose a deep learning based semantic segmentation algorithm to identify and label the tissues and organs in the endoscopic video feed of the human torso region.
\absdiv{Methods}
We present an annotated dataset, m2caiSeg, created from endoscopic video feeds of real-world surgical procedures. Overall, the data consists of 307 images, each of which is annotated for the organs and different surgical instruments present in the scene. We propose and train a deep convolutional neural network for the semantic segmentation task. To cater for the low quantity of annotated data, we use unsupervised pre-training and data augmentation.
\absdiv{Results}
The trained model is evaluated on an independent test set of the proposed dataset. We obtained a F1 score of 0.33 while using all the labeled categories for the semantic segmentation task. Secondly, we labeled all instruments into an 'Instruments' superclass to evaluate the model's performance on discerning the various organs and obtained a F1 score of 0.57. 
\absdiv{Conclusion}
We propose a new dataset and a deep learning method for pixel level identification of various organs and instruments in a endoscopic surgical scene. Surgical scene understanding is one of the first steps towards automating surgical procedures.
\end{abstract}


\section{Introduction}
\label{sec:introduction}
Deep Learning has revolutionized a diverse set of domains ranging from Computer Vision to Natural Language Processing to Reinforcement Learning. These in turn have application from self driving cars to healthcare to document analysis to advancement of Artificial General Intelligence. Just like large scale datasets, like ImageNet \cite{deng2009imagenet}, brought about significant advances in image classification, datasets like the PASCAL VOC \cite{everingham2010pascal} and MS COCO \cite{lin2014microsoft} have made possible training of powerful Deep Learning models for Object Detection and Semantic Segmentation. Especially considering the domain of autonomous vehicles, datasets like KITTI \cite{geiger2013vision} and CamVid \cite{brostow2009semantic} combined with the state of art Deep Neural Networks have made large strides possible. Self-driving cars, powered by Deep Learning, are predicted to be in production and use within the next few years. We expect that Artificial Intelligence (AI) will similarly bring significant advances in healthcare and surgical procedures as well. To this end, we propose semantic segmentation of laparoscopic surgical images as a first step towards autonomous robotic surgeries.

Deep Learning has already made possible many useful methods in healthcare, including predicting diabetic retinopathy \cite{gulshan2016development} and cardiovascular risk factors \cite{poplin2017predicting} from retinal images, and breast cancer detection \cite{liu2017detecting} from pathological images. However, very little work has been done in the domain of automating robotic surgical procedures. While complex systems, like the DaVinci surgical robot, have made surgical operations more precise and quicker, lead to less blood loss than conventional surgeries, and lead to quicker patient recovery times; they require a high level of skill and domain knowledge.

 Recent advances in AI and Robotics can, however, be used to automate such procedures, which can result in even more precise procedures (removing human error), and effective utilization of doctors and medical practitioners time so that they can attend to other pressing issues and medical research. Our proposed work aims to set the foundation for such work.
 
 Previous work combining Semantic Segmentation and Robotic surgeries has been limited to instrument segmentation. The usage of deep learning has been limited to Tool Presence Detection (multi-class classification) \cite{twinanda2017endonet, raju2016m2cai} and Surgical Phase Identification \cite{volkov2017machine, stauder2016tum, jin2016endorcn, twinanda2017endonet}. However, these existing works do not facilitate automating surgical procedures. In particular, at the very least, we need to precisely identify the different organs, instruments, and other entities present in surgical images and videos. Once we have a good understanding of the presented scene, only then can we think of automating the procedure. Our work presents the problem where we have to identify, at a pixel level, every entity present in images of such procedures.
 
 The main contributions of this paper are:
 
 \begin{enumerate}
 	\item Proposal of the surgical scene segmentation problem as one of the first steps towards autonomous surgical procedures. 
 	\item A novel dataset, m2caiSeg, which can be used to train and evaluate any proposed algorithms for the task.
 	\item An Encoder-Decoder Convolutional Neural Network architecture for addressing the problem.
 	\item Baseline results which the research community can build upon and improve.
 	\item Open-source code and dataset.
 \end{enumerate}
 
 The rest of the paper is organized as follows: Section \ref{sec:literature_review} gives an overview of the existing literature in the domain. Section \ref{sec:proposed_architecture} discusses the proposed dataset and neural network architecture. In Section \ref{sec:results_and_evaluation}, we present the results of our network on the dataset. Finally, in Section \ref{sec:conclusion}, we conclude the paper and present some potential directions for future work.
 
\section{Literature Review}
\label{sec:literature_review}
We present the existing work related to the Laparoscopic Image and Video Analysis, and Semantic Segmentation.

\subsection{Laparoscopic Image and Video Analysis}
\label{sub:laparoscopic_image_and_video_analysis}
Prior work done in Laparoscopic image and video analysis focuses primarily on three different aspects:

\begin{enumerate}
	\item Surgical Phase Segmentation
	\item Tool Presence Detection
	\item Surgical Tool Segmentation
\end{enumerate}

We briefly discuss these domains in the following subsections.

\subsubsection{Surgical Phase Identification}
\label{subsub:surgical_phase_identification}
Surgical phase identification (also referred to as surgical phase segmentation) refers to the identification of the temporal phase of a surgical procedure. A surgical operation is sub-categorized into different phases of the surgery. This has applications in surgical coaching, education, automated and assisted surgical procedures, and post surgical analysis of the operation.

Recently, Volkov et al. \cite{volkov2017machine} proposed a method which uses color, organ position, shape (for instruments), and texture features to obtain a Bag-of-Words (BOW) representation of frames in surgical videos. They use multiple binary Support Vector Machine (SVM) classifiers for each phase to classify frames. They then use a temporal HMM to correct the initial SVM predictions. The videos of the Laparoscopic Vertical Sleeve Gastrectomy procedure were used and segmented into seven distinct phases. The authors obtained 90.4 \% accuracy using SVMs, and raised it to 92.8 \% with the HMM correction.

The M2CAI 2016 Surgical Workflow challenge held as part of the Medical Image Computing and Computer Aided Intervention (MICCAI) conference in 2016 introduced the TUM LapChole dataset \cite{stauder2016tum}. The dataset contains 20 videos (15 Training and 5 Test) of the Laparoscopic Cholecystectomy procedure. The videos are annotated and categorized into 8 distinct phases. They also provided baseline results for the challenge by using an AlexNet \cite{krizhevsky2012imagenet} model trained and tested on 1 frame extracted per second. They later used a sliding window approach to correct misclassifications by taking the majority vote among the last 10 predictions. The baseline results were average Jaccard Index, average Precision, and average Recall of 52.4 \%, 65.9 \%, and 74.7 \% respectively. The challenge winning entry from Jin et al. \cite{jin2016endorcn} used a Recurrent Convolutional Network model, EndoRCN. They used a 50 layer ResNet \cite{he2016deep} trained for classification into eight categories as a visual feature extractor. Secondly, they used the current frame and the previous 2 frames to extract the visual features using the ResNet model. The 3 extracted features were fed sequentially to a LSTM model, which was used to predict the phase. Post-processing using sequential consistency was performed to further improve the predictions. The authors achieved a Jaccard Index score of 78.2.

Another important work in this domain was proposed by Twinanda et al. \cite{twinanda2017endonet}, where they introduced another dataset, Cholec80. The Cholec80 dataset contains 80 videos of the Cholecystectomy procedure, sampled at 1 FPS, where each frame is annotated with the surgical phase information, and additionally, also the tool presence annotations. The surgical phases are divided into 8 distinct categories, while there are tool presence annotations for 7 different surgical instruments. The authors use a modified AlexNet \cite{krizhevsky2012imagenet} architecture, which predicts both tool presence in a frame, and uses that, along with the network features, to predict the surgical phase.

\subsubsection{Tool Presence Detection}
\label{subsub:tool_presence_detection}
Tool Presence Detection is a multi-class multi-label classification problem, where a mapping is desired from image pixels to a vector representing the presence of surgical tools in the image. This problem can be framed as an image classification problem: a field Machine Learning and Convolutional Neural Networks has dominated in recent years. Tool Presence Detection has applications in automated and assistive surgeries, surgical workflow analysis, and as \cite{twinanda2017endonet} showed, in aiding phase segmentation.

As discussed earlier in Section \ref{subsub:surgical_phase_identification}, the EndoNet architecture \cite{twinanda2017endonet} was used for joint training of tool presence detection and surgical phase segmentation. This particular work also led to the M2CAI Surgical Tool Detection Challenge held at MICCAI 2016. The challenge dataset, hereby referred to as M2CAI-tool dataset, consists of 15 videos of cholecystectomy procedures, of which there are 10 training videos and 5 test videos. The dataset contains tool presence annotation of the following 7 tools:

\begin{itemize}
	\item Grasper: \textit{Used to grasp and maneuver the different organs and tissues}
	\item Bipolar: \textit{Used to seal tissues to stop haemorrhages or blood loss}
	\item Hook: \textit{Used to burn tissue for ease of later dissection}
	\item Clipper: \textit{Used to seal tissues and blood vessels before dissection}
	\item Scissors: \textit{Used for tissue dissection}
	\item Irrigator: \textit{Used to introduce saline water in case of bleeding/bile. Also used as a fluid suction}
	\item Specimen Bag: \textit{Used to collect and bring the dissected organ out of the body}
\end{itemize}

Not surprisingly, the challenge winning entry by Raju et al. \cite{raju2016m2cai} used Convolutional Neural Networks for the task. They used an ensemble of the popular VGGNet \cite{simonyan2014very} and GoogLeNet \cite{szegedy2015going} architectures to achieve a mean Average Precision (mAP) of 63.7 \% on the 5 videos in the test set. Our work builds upon the M2CAI-tool dataset for semantic segmentation.

\subsubsection{Surgical Tool Segmentation}
\label{subsub:surgical_tool_segmentation}
Surgical Tool Segmentation is identifying the location at the pixel level in an image, where the surgical tool(s) lie. Surgical Tool Segmentation is one of the important research areas explored since a few years in the domain of computer-assisted surgical systems. This is important since it can provide feedback and other guidance signals to the surgeon. This also helps immensely in surgeries requiring higher precision. Segmentation of the tool at the pixel level is important because of the critical nature of surgical procedures. Accurate tool segmentation can then lead to accurate tool localization and tracking. This step is also essential for automating surgeries, but we also need information about non-instrument part of the scene. Nevertheless, since our work is an extension of this, we would first like to discuss some approaches for the task.

Traditionally, image processing techniques were the dominant approach for Surgical Tool Segmentation. Since the scenario has changed with the success of Deep Learning algorithms, we would focus more on those. But we would describe one of the methods for reference. Doignon et al. \cite{doignon2005real} introduced a method based on a combination of various image processing techniques, including the use of hue, saturation, edge detection, region growing, and using shape features to classify regions in an image.

More recently though, Garc{\'\i}a-Peraza-Herrera et al. \cite{garcia2016real} proposed a real-time tool segmentation method, which uses Fully Convolutional Networks (FCNs) \cite{long2015fully} along with Optical Flow based tracking to segment surgical instruments in videos. Due to hardware limitations of running FCN inference in real-time, they use OpticalFlow tracking and assuming somewhat rigidity of the tool and scene for a few frames, they compute affine transformation of the new segmentation mask with respect to the previous one. The segmentation masks are updated as the FCN computes the results; enabling real-time segmentation. However, with today's hardware and efficient Deep Learning architectures, using a purely Deep Learning system for real-time segmentation is very much possible. More recently, Garc{\'\i}a-Peraza-Herrera et al. \cite{garcia2017toolnet} introduced ToolNet, a modified version of the FCN. They introduced 2 different architectures, one which aggregated predictions across multiple scales before calculating the loss, and another which incorporated a multi-scale loss function. They also used the Dice Loss \cite{milletari2016v, li2017compactness}, which has shown to be effective for semantic segmentation, especially across unbalanced classes. We also incorporate the Dice Loss in our work. They also used Parametric Rectified Linear Units (PReLU) \cite{he2015delving}, an adaptive version of the Rectified Linear Unit (ReLU) as the activation function. They achieve a Balanced Accuracy of 81.0 \% and a mean Intersection over Union (IoU) of 74.4 \% on the Da Vinci Robotic (DVR) dataset, which was part of the MICCAI 2015 Endoscopic Vision (EndoVis) Challenge.

Attia et al. \cite{attia2017surgical} proposed a Hybrid CNN-RNN Encoder-Decoder network for surgical tool segmentation. They used a 7 convolution layered CNN to extract feature maps from the input image. Using just an Encoder-Decoder network produced coarse segmentation masks. To cater for this and to account for spatial dependencies among neighboring pixels and to enforce global spatial consistency, the authors used 4 Long Short Term Memory (LSTM) Recurrent Neural Network (RNN) layers in sequence on the produced encoder feature maps. They then used a decoder network to upsample the feature maps into the final segmentation masks. They achieved a balanced accuracy of 93.3 \% and an IoU of 82.7 \% with their method on the MICCAI 2016 Endoscopic Vision (EndoVis) Challenge Robotic Instruments dataset.

Another important advancement in surgical tool segmentation was the segmentation of the tool into it's constituent parts, i.e., the tool shaft and the tool's manipulator. This transforms the binary classification/segmentation problem into a 3-class classification problem, where the 3rd category is Background. Pakhomov et al. \cite{pakhomov2017deep} proposed a 101-layered ResNet \cite{he2016deep} model which is casted as a FCN for semantic segmentation. They use Dilated Convolutions \cite{yu2015multi} to reduce the down-sampling induced by Convolutional layers without padding. Dilated (also called Atrous) convolutions have proven useful for semantic segmentation as it allows larger receptive fields while keeping the number of network parameters low. They achieved the state of the art results (at the time of their paper) on the MICCAI 2015 EndoVis Challenge Robotic Instruments dataset at 92.3 \% Balanced Accuracy for Binary Classification/Segmentation.

Since instrument segmentation and localization are interdependent tasks, Laina et al. \cite{laina2017concurrent} utilize that for concurrent segmentation and localization of surgical instruments. Additionally, they frame localization as a heatmap regression problem, where they use landmark points on the instruments with a Gaussian centered around them to generate the groundtruth. They then regress for the heatmaps (one per landmark), which represent the confidence of each pixel to be in the proximity of the groundtruth landmark. This approach makes training easier and stable over regressing over 2D (x, y) coordinates of the landmark points. The authors train jointly for both segmentation and localization, which helps in improving performance for both tasks. They also use a multi-class segmentation approach, similar to \cite{pakhomov2017deep}, but with 5 different classes namely: Left shaft, Right shaft, Left Grasper, Right Grasper, Background. They obtain a Balanced Accuracy of 92.6 \% on the MICCAI 2015 EndoVis Challenge.

As we can see from the above approaches, Deep Neural Networks, and CNNs in particular, have been the de facto approach for various tasks in Laprascopic Image and Video analysis in the previous few years. And rightly so since their success in this domain builds upon the success of Deep Learning in general over the last few years. It should also be noted that joint training for multiple objectives has shown to be helpful in training for all the objectives.

That said, all the above methods focus on instrument segmentation. For a better and more complete understanding of a robotic scene, especially considering autonomous robotic surgeries, this is insufficient. We need to know the precise location of not just the tools, but also the organs. Till date, to the best of our knowledge, no such approach for dense instrument and organ segmentation has been explored in literature.


\subsection{Semantic Segmentation}
\label{sub:semantic_segmentation}
Semantic segmentation is the pixel level labeling/classification for any image/video. It is the natural step after success of several deep learning based object detection networks \cite{girshick2014rich, girshick2015fast, ren2015faster, redmon2016you, redmon2017yolo9000, liu2016ssd}, where objects are located by a bounding box. Object detection at pixel level, or getting an accurate object mask, is critical for many applications such as self-driving cars, and especially in our case of robotic surgeries.

Long et al. \cite{long2015fully} introduced the first popular end-to-end trainable Deep Learning architecture for Semantic Segmentation, the Fully Convolutional Network (FCN). FCN is a deep CNN which uses a series of Convolution and Pooling layers to generate feature maps. The feature maps are then upsampled using Fractionally-Strided Convolutions. Fractionally-Strided Convolutions or Transpose Convolutions (also sometimes wrongly referred to as Deconvolution) zero-pad the input feature map between the pixels, where the number of zeros is the scale factor (k) - 1. A regular convolution is then performed on this fractionally padded input. This results in a learnable and differentiable upsampling filter. In addition to Fractionally Strided Convolutions, the authors combine information across different layers (which results in information sharing across different scales). Lastly, they use an \textit{n}-way softmax for each pixel for prediction, where \textit{n} is the number of classes.

Soon after the release of FCN, Badrinarayanan et al. \cite{badrinarayanan2015segnet} proposed an Encoder-Decoder network for Semantic Segmentation called SegNet. The Encoder part of the network is identical to the VGG network \cite{simonyan2014very}, where the max pooling indices for each layer are stored for upsampling later. In the Decoder part of the network, feature maps are successively upsampled using the corresponding max pooling indices into sparse feature maps. These sparse feature maps are then convolved with learnable filters to obtain dense feature maps, and ultimately a semantic segmentation of the input.

Another Semantic Segmentation architecture worth discussing is the U-Net architecture \cite{ronneberger2015unet}, which was proposed for biomedical image segmentation. The main feature of the proposed U-Net architecture is feature concatenation (or information sharing) from earlier in the network to the later layers. This helps retain low level features, like edges, which helps in obtaining sharper segmentation masks. The architecture has been successful and popular for biomedical image segmentation tasks.

Most popular (and powerful) Semantic Segmentation models use Conditonal Random Fields (CRFs) \cite{lafferty2001conditional} to post-process and refine the network predictions. This makes the network step-wise and not end to end trainable. To circumvent this, Zheng et al. \cite{zheng2015conditional} introduced the CRF-RNN model, where they modeled the CRF component of the network as a Recurrent Neural Network. This in turn resulted in an end-to-end trainable network capable of semantic segmentation with a trainable CRF component.

More recently, another popular architecture, the Fully Convolutional DenseNet \cite{jegou2017one} was proposed by J{\'e}gou et al. This architecture tries to remove the CRF out of the equation by modifying the DenseNet architecure \cite{huang2017densely} for Semantic Segmentation. \cite{huang2017densely} proposed the DenseNet, architecture which is composed of Dense blocks (which can be said as an extension to the Residual Blocks in \cite{he2016deep}). The dense blocks function by concatenating the outputs of all previous layers (including the first input) with the output of the current layer. The problem with such an architecture, especially for semantic segmentation as it requires a Decoder network as well, is the explosion in the number of feature maps as we go deeper in the network. To cater to this problem, \cite{jegou2017one} do not concatenate the Dense Block input to the output in the Decoder Network. They also use skip connections to combine information from the Encoder network with the Decoder network.

\section{Proposed Architecture}
\label{sec:proposed_architecture}
Our work builds upon the methods for Laparoscopic Image and Video Analysis and the general success of Convoluional Neural Networks for Semantic Segmentation, as discussed in Section \ref{sec:literature_review}. Since no dataset for the task existed at the initiation of the project, we made and annotated a dataset for semantic segmentation of robotic surgical scenes. Section \ref{sub:the_m2caiseg_dataset} describes our work on the m2caiSeg dataset, while Section \ref{sub:proposed_network_architecture} presents our proposed network architecture.

\subsection{The m2caiSeg dataset}
\label{sub:the_m2caiseg_dataset}
Our proposed m2caiSeg dataset is an extension of a small subset of MICCAI 2016 Surgical Tool Detection dataset \cite{twinanda2017endonet} (M2CAI-tool). The M2CAI-tool dataset, described briefly earlier in Section \ref{subsub:tool_presence_detection}, consists of a total of 15 videos, which are divided into 10 training videos and 5 test videos. Each video has a tool presence annotation every 25 frames, i.e., at a rate of 1 FPS. There are a total of 7 tools as detailed in Section \ref{subsub:tool_presence_detection}.

\subsubsection{Annotation Methodology}
\label{subsub:annotation_methodology}
We developed a tool in MatLab, based on MegaPixels\cite{sajid2016robust}, which segments the region into distinct parts. However, sometimes the tool produced regions which were not representative of the true segmentation boundaries. For such images, we had to manually annotate the difficult regions or region boundaries using Microsoft Paint or Photoshop first before passing it to the annotation tool. However, using the tool did help us in reducing the annotation time.

\subsubsection{Dataset Details}
\label{subsub:dataset_details}
We sub-sampled 307 images from Videos 1 and 2 of the M2CAI-tool training set and annotated them at a pixel level into various different categories and sub-categories as shown below:

\begin{itemize}
	\item \textbf{Organs:} Liver, Gallbladder, Upperwall, Intestine
	\item \textbf{Instruments:} Grasper, Bipolar, Hook, Scissors, Clipper, Irrigator, Specimen Bag, Trocars (Provide an opening to insert the surgical instruments), Clip (The clips applied by the Clipper to seal the blood vessels)
	\item \textbf{Fluids:} Bile, Blood
	\item \textbf{Miscellaneous:} Unknown (Used as a label for pixels which are indiscernible for the annotator), Black (Used as a label for the surrounding region in the image which is not visible due to the trocar limiting the camera field of view)
	\item \textbf{Artery}
\end{itemize}

\begin{figure}
	\centering
	\includegraphics[width=0.75\textwidth]{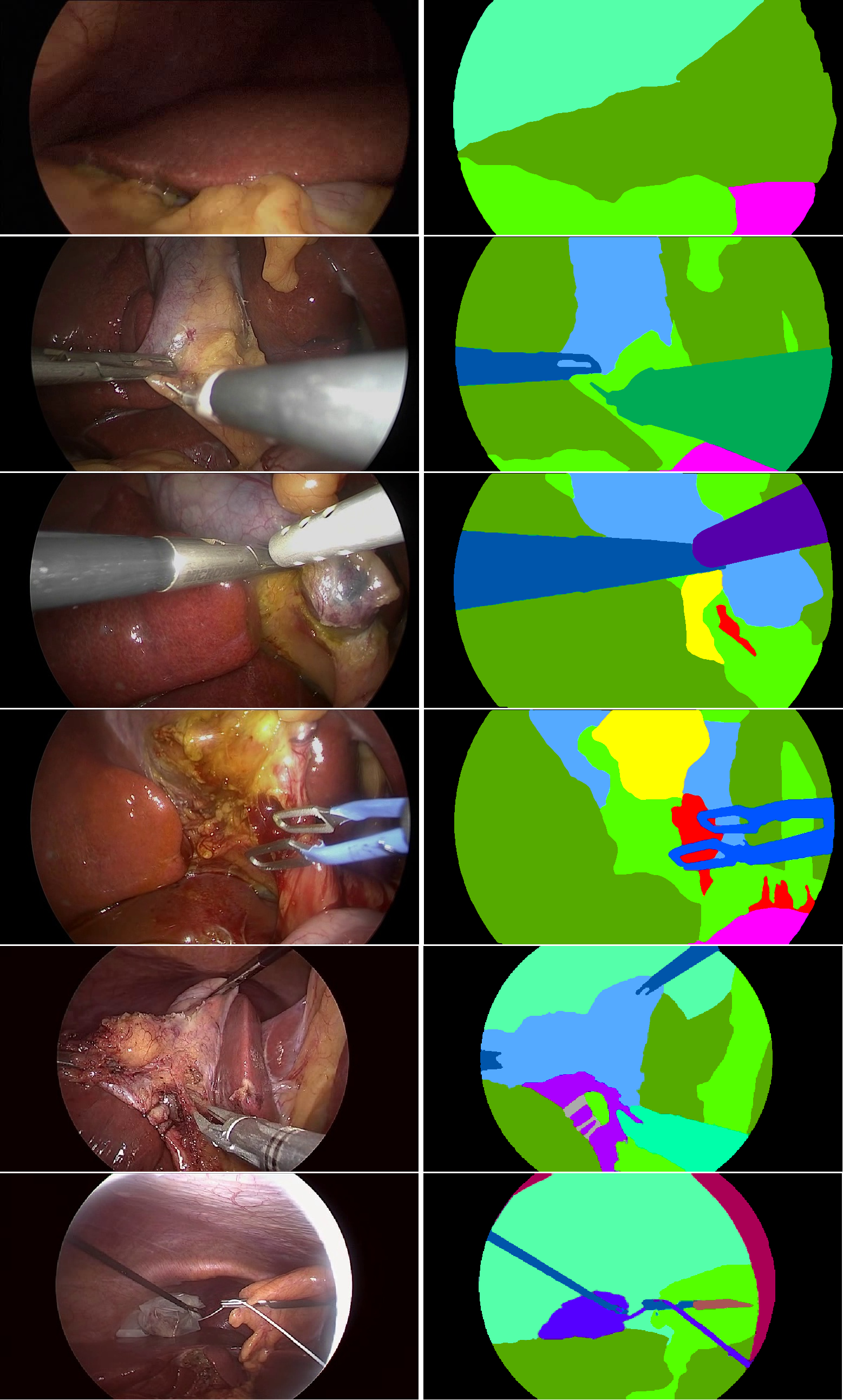}
	\caption{Some samples from the m2caiSeg dataset. The left column shows the original images while the right column shows their corresponding groundtruth annotations.}
	\label{fig:m2caiseg_samples}
\end{figure}

Altogether, we annotated a total of 5 different organs, 9 different instruments (of which Trocars and Clip are different from the Tool presence annotation in M2CAI-tool dataset), 2 fluids, 2 miscellaneous categories, and the artery. The annotations were done while considering the usage of this information for autonomous robotic surgeries. For example, in this context, the Clip needs to be identified in the videos to be able to learn where and when to place it. Similarly, the artery needs to be identified since it needs to be sealed. Blood loss indicates potential use of the bipolar to seal any open incisions, while Bile indicates that the irrigator potentially needs to be used for sterilization and cleanup. The other categories are self-explanatory. Figure \ref{fig:m2caiseg_samples} shows some sample annotations from our dataset.

\subsection{Proposed Network Architecture}
\label{sub:proposed_network_architecture}
\begin{figure}
	\centering
	\includegraphics[width=\textwidth]{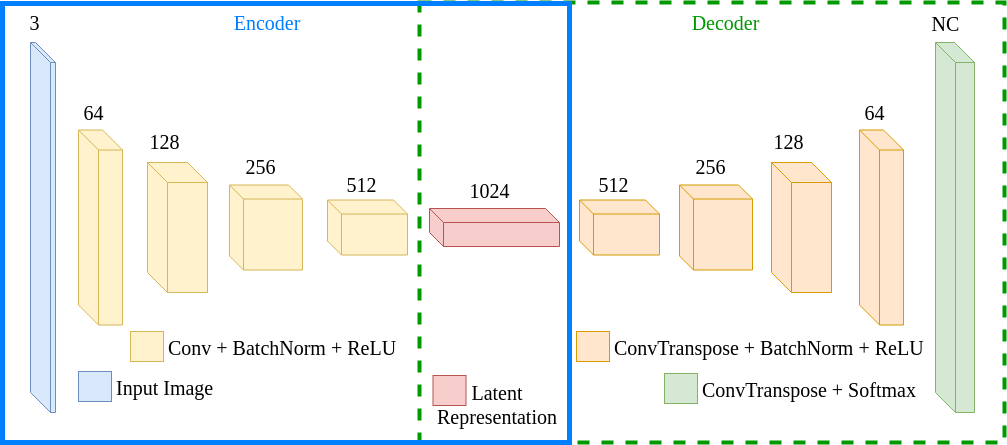}
	\caption{Our proposed 10 layer CNN Encoder-Decoder Network. NC means the number of classes, which is equal to the channel dimensions of the network output.}
	\label{fig:proposed_network}
\end{figure}

Due to the relatively small quantity of annotated data, we focused our efforts on methods which are data efficient or which work well with smaller datasets like ours. We propose a minimalist Encoder-Decoder Convolutional Neural Network as shown in Figure \ref{fig:proposed_network}. We will refer to this as the segmentation network.

The input image is resized to 256 x 256 at training time. Due to the small size of our dataset, we perform online 10-crop data augmentation at training time, where we take all 224 x 224 crops from all the 4 corners and the centre, and their horizontal flips (mirror images). We normalize each image by its RGB per-channel mean [0.295, 0.204, 0.197] and standard deviation [0.221, 0.188, 0.182]. These values are computed over all the 581923 frames of the M2CAI-tool training set. At test time, the input image is normalized using the same mean and standard deviation, but we do not use 10-crop data augmentation. In this case, the input image is resized to 256 x 256, but since we do not take any crops, the resolution stays at 256 x 256 and does not go to 224 x 224. We used convolution layers with 64, 128, 256, 512, and 1024 filters, respectively, for the Encoder part to obtain the latent representation. We used a kernel size of 4 x 4, stride of 2, and padding of 1 for all the convolution layers in our network except the last encoder layer where we used the same kernel size but with a stride of 1 and no padding. Convolution layers are followed by batch normalization \cite{ioffe2015batch} and the ReLU non-linearity. It is important to note that we do not use batch normalization in the first Encoder layer in our network.

Once we have obtained the latent representation of our input image after it passes through the Encoder, it then goes through the Decoder network where it is successively upsampled using Convolution Transpose Layers based on Fractionally Strided Convolutions. We mirror the network about the Latent representation, and thus our Convolution Transpose layers use the same hyperparameters as the convolution layers in the Encoder network, i.e., the first Decoder layer uses a stride of 1 and no padding, while all the other layers use stride of 2 and padding of 1. The kernel size is 4 x 4 for all the layers. The number of filters used in Convolution Transpose layers are 512, 256, 128, and 64, successively, while the last layer uses filters equal to the number of classes. We also use BatchNorm and the ReLU non-linearity in all Convolution Transpose layers except the last one, where we use a softmax for each pixel. The network output is the same size as the resized input image, i.e., 224 x 224 and 256 x 256 for training and inference, respectively. We also use 2D Dropout \cite{srivastava2014dropout} in the first three Decoder layers of our network with a dropout probability of 0.5. Finally, a pixel-wise softmax converts the Decoder output to class-wise probabilities for each pixel.

\subsection{Training Details}
\label{sub:training_details}
We split our dataset into a training set and a test set, containing 245 images and 62 images, respectively. Since our dataset size is rather limited, we explored unsupervised pre-training to learn dataset specific features. We believe this can be particularly helpful for Semantic Segmentation.

We train over the entire M2CAI-tool training set (581935 frames) for image reconstruction over 1 epoch. The network used is the same as the segmentation network, except that the last layer of the network uses a Sigmoid instead of a Softmax. We used a learning rate of 0.01, and a batch size of 64 for the training. For the loss function, we used the per-pixel Mean Squared Error loss, which is given as:

\begin{equation}
MSE = \frac{1}{n} \sum_{i = 1}^{n} (y - \hat{y})^2
\end{equation}

\noindent where \textit{n} is the number of pixels, \textit{y} is the groundtruth pixel RGB value, and \textit{$\hat{y}$} is the predicted pixel RGB value. We initialize our segmentation network with weights from the reconstruction network. We then finetune the network for semantic segmentation. We use the Adam Optimizer \cite{kingma2014adam}  to train both our reconstruction and segmentation networks with $\beta\textsubscript{1}$ and $\beta\textsubscript{2}$ as 0.9 and 0.999, respectively. We also used Weight Decay as a regularizer with $\lambda$ equals 0.0005. We used step learning rate decay with the initial learning rate 0.0001, which is halved every 10 epochs. We trained for a total of 90 epochs with a batch size of 2, which was chosen based on the hardware specifications.

We used a multi-class pixel-wise Dice loss function. The Dice Similarity Coefficient (DSC) measures the similarity between two image regions. However, in its discrete form, the DSC is not differentiable and thus can not be used as a loss function. \cite{milletari2016v} proposed a continuous and differentiable version of the DSC, which can be used directly as a loss function in training Deep Neural Networks. We use that formulation as the loss function of our network. Using the Dice Loss function leads to a better evaluation accuracy in our case as compared to the Cross-Entropy loss.

\section{Results and Evaluation}
\label{sec:results_and_evaluation}
We evaluate our proposed network on the m2caiSeg dataset proposed in Section \ref{sec:proposed_architecture}. We divide our training and evaluation into two different categories:

\begin{enumerate}
	\item Single Instrument Class: \textit{We categorize all instruments into one single class, i.e., Instruments}
	\item All Categories: \textit{We use all categories as outlined in Section \ref{subsub:dataset_details}}
\end{enumerate}

For evaluation, we report 4 different performance measures, namely Intersection over Union (IoU) (also called Jaccard Index), Precision, Recall, and the F1 score for each class, as well as their mean over all classes.

\begin{equation}
IoU = \frac{TP}{TP + FP + FN}
\end{equation}

\begin{equation}
Precision = \frac{TP}{TP + FP}
\end{equation}

\begin{equation}
Recall = \frac{TP}{TP + FN}
\end{equation}

\begin{equation}
F1 Score = \frac{2 * Precision * Recall}{Precision + Recall}
\end{equation}

We use a pixel-wise criteria to evaluate the True Positives (TP), False Positives (FP), and the False Negatives (FN). The IoU represents the degree of overlap between the segmentation regions, benefiting from the True Positives while penalizing both the False Positives and False Negatives. Precision and Recall represent the resilience to False Positives and False Negatives respectively. Finally, the F1 score is the harmonic mean between Precision and Recall and gives a more balanced estimate taking into account both the False Positives and False Negatives.

\subsection{Single Instrument Class}
\label{sub:single_instrument_class}
These experiments use a subset of all 19 classes, clustering all the instruments into 1 super-category, and merging the Fluid super-category (Blood and Bile) with the Gallbladder class; resulting in a total of 9 classes. Table \ref{tab:single_instrument_class_results} shows the categories and the results of our proposed network on the mentioned categories.

\begin{table}[!htb]
	\centering
	\begin{tabular}{|a|p{1.2cm}|p{1.2cm}|p{1.2cm}|p{1.2cm}|p{1.2cm}|p{1.2cm}|p{1.2cm}|p{1.2cm}|}
		\hline
		\rowcolor{Gray}
		{Class} & {IoU (Proposed Method)} & {IoU (UNet)} & {Precision (Proposed Method)} & {Precision (UNet)} &  {Recall (Proposed Method)} & {Recall (UNet)} & {F1 Score (Proposed Method)} & {F1 Score (UNet)}\\
		\hline
		Unknown & 0.00 & 0.00 & 0.00 & 0.01 & 0.00 & 0.00 & 0.00 & 0.00 \\
		Instruments & 0.73 & 0.51 & 0.79 & 0.68 & 0.91 & 0.67 & 0.85 & 0.68 \\
		Liver & 0.77 & 0.54 & 0.84 & 0.61 & 0.90 & 0.84 & 0.87 & 0.70 \\
		Gallbladder & 0.50 & 0.19 & 0.80 & 0.40 & 0.58 & 0.26 & 0.67 & 0.31 \\
		Fat & 0.53 & 0.39 & 0.61 & 0.69 & 0.79 & 0.47 & 0.69 & 0.56 \\
		Upper Wall & 0.41 & 0.08 & 0.65 & 0.15 & 0.53 & 0.14 & 0.58 & 0.14 \\
		Intestine & 0.17 & 0.00 & 0.80 & 0.00 & 0.18 & 0.00 & 0.30 & 0.00 \\
		Artery & 0.09 & 0.00 & 0.49 & 0.00 & 0.09 & 0.00 & 0.16 & 0.00 \\
		Black & 0.94 & 0.90 & 0.96 & 0.92 & 0.97 & 0.94 & 0.97 & 0.93 \\
		\textbf{Mean} & 0.46 & 0.29 & 0.66 & 0.38 & 0.55 & 0.37 & 0.57 & 0.37 \\ \hline
	\end{tabular}
	\caption{Results of our network with pre-training for image reconstruction on the m2caiSeg dataset with a single instrument class. The table also compares our network results with U-Net \cite{ronneberger2015unet}, a popular architecture for Biomedical Image Segmentation.}
	\label{tab:single_instrument_class_results}
\end{table}

As we can see from Table \ref{tab:single_instrument_class_results}, our network performs well on the majority classes in the dataset, while its performance suffers on the less dominant classes; especially Intestine and Artery. This is expected since the number of instances of these two classes are quite low. This is also the case for the Unknown class as our network fails to correctly predict that class throughout the evaluation. This can potentially be explained by the fact that while human annotators did not get how a particular part of the image should be labeled, and hence annotated it as Unknown; the algorithm learns features through which it is able to make a prediction other than Unknown for those image regions.

\begin{figure}
	\centering
	\includegraphics[width=0.9\textwidth]{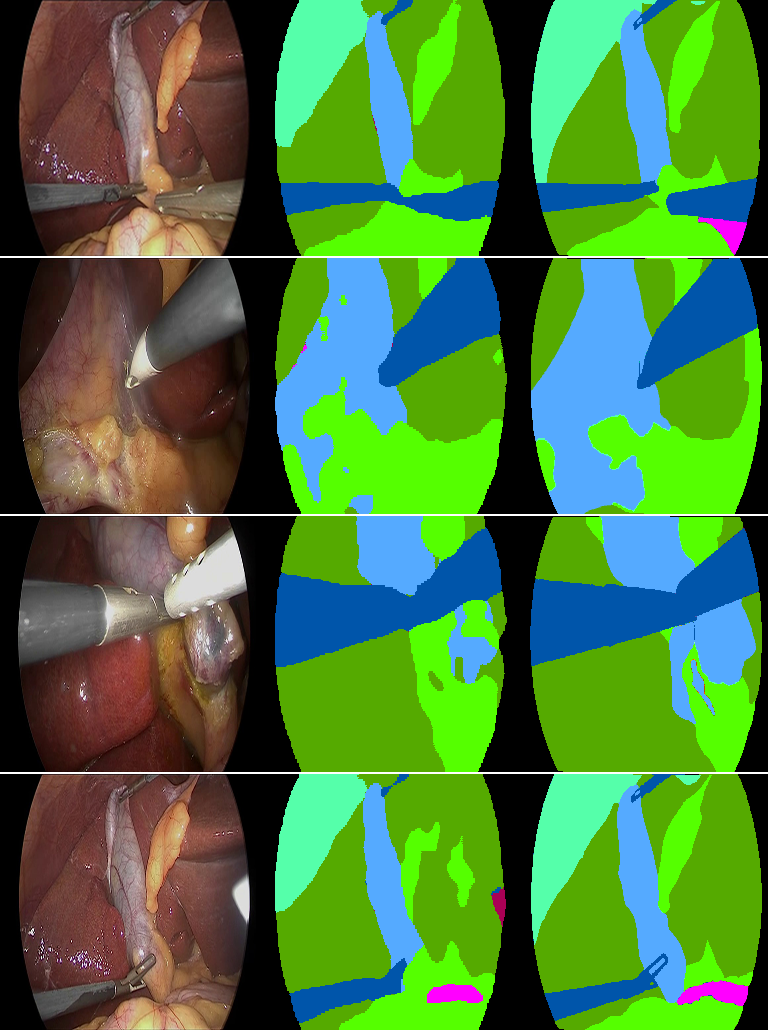}
	\caption{Predictions of our network on the m2caiSeg dataset with the single instrument class. The left column shows the original images, the middle column shows the prediction, while the right column shows the corresponding groundtruth.}
	\label{fig:single_instrument_class_predictions}
\end{figure}

\begin{figure}
	\centering
	\includegraphics[width=0.9\textwidth]{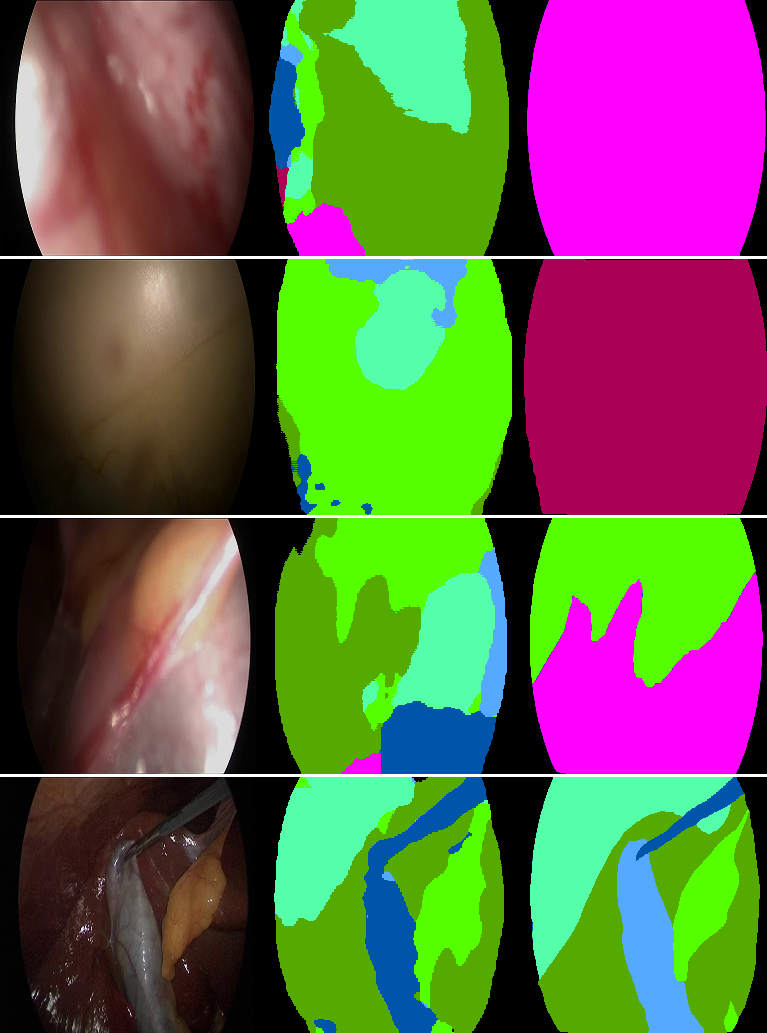}
	\caption{Some failure cases of our network on the m2caiSeg dataset with the single instrument class. The left column shows the original images, the middle column shows the prediction, while the right column shows the corresponding groundtruth.}
	\label{fig:single_instrument_class_failures}
\end{figure}

That being said, the more dominant classes especially Instruments, Liver, and Black perform well. However, the algorithm fails to impress for Gallbladder and Fat classes, which are also common. The algorithm mostly confused the Gallbladder and Instruments class. Figure \ref{fig:single_instrument_class_predictions} shows some of the predictions for our network, while Figure \ref{fig:single_instrument_class_failures} shows some failure cases for our network.

As can be seen from Figure \ref{fig:single_instrument_class_failures}, most failures are for images which are difficult to discern. Additionally, our network sometimes confuses the Gallbladder for an Instrument due to potentially similar colors and shape.

U-Net \cite{ronneberger2015unet}, as discussed earlier in Section \ref{sec:literature_review}, is a popular architecture for Biomedical Image Segmentation. We compare the performance of our network with U-Net. The U-Net architecture was trained from scratch, with the same hyperparameters as our proposed network. The results are presented in Table \ref{tab:single_instrument_class_results}.

\begin{figure}
	\centering
	\includegraphics[width=\textwidth]{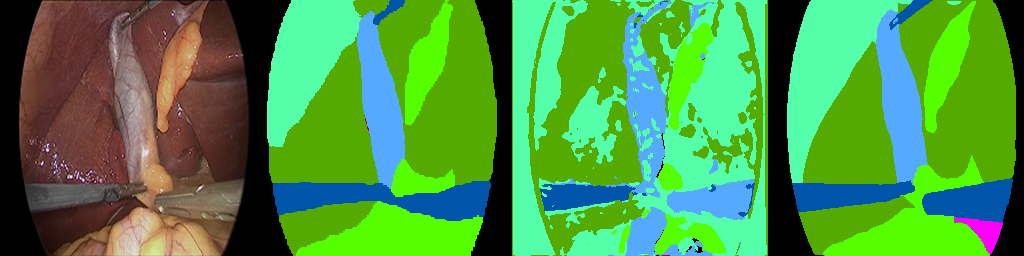}
	\caption{Sample of predictions of our network as compared to UNet \cite{ronneberger2015unet}, a popular CNN architecture for Biomedical image segmentation. The left column shows the input image, the 2nd column shows the prediction for our network, the 3rd column shows prediction with UNet, and the last column shows the groundtruth.}
	\label{fig:comparison_with_unet}
\end{figure}

Our proposed method outperforms UNet in all categories of all evaluation criteria. It also shows that the unsupervised pre-training is especially beneficial for small datasets in Semantic Segmentation. Figure \ref{fig:comparison_with_unet} shows the result for our proposed method and UNet for a particular image from our test set.

As we can see from Figure \ref{fig:comparison_with_unet}, UNet preserves the lower level features nicely, such as edges and shape information, but fails to make fine predictions globally. We propose some interesting directions in Section \ref{sec:conclusion} on how to make use of this information.

\subsection{All Categories}
\label{sub:all_categories}
We additionally perform experiments for training and evaluation on all 19 classes of the m2caiSeg dataset. The results are summarized in Table \ref{tab:all_categories_results}.

\begin{table}[!htb]
	\centering
	\begin{tabular}{abbbb}
		\hline
		\rowcolor{Gray}
		\textsc{Class} & \textsc{IoU} & \textsc{Precision} & \textsc{Recall} & \textsc{F1 Score}\\
		\hline
		Unknown & 0.00 & 0.00 & 0.00 & 0.00 \\
		Grasper & 0.39 & 0.64 & 0.49 & 0.56 \\
		Bipolar & 0.00 & 0.00 & 0.00 & 0.00 \\
		Hook & 0.64 & 0.70 & 0.89 & 0.78 \\
		Scissors & 0.00 & 0.00 & 0.00 & 0.00 \\
		Clipper & 0.34 & 0.51 & 0.51 & 0.51 \\
		Irrigator & 0.01 & 0.71 & 0.01 & 0.02 \\
		Specimen Bag & 0.22 & 0.50 & 0.29 & 0.37 \\
		Trocars & 0.00 & 0.00 & 0.00 & 0.00 \\
		Clip & 0.00 & 0.00 & 0.00 & 0.00 \\
		Liver & 0.73 & 0.78 & 0.92 & 0.85 \\
		Gallbladder & 0.53 & 0.71 & 0.68 & 0.70 \\
		Fat & 0.55 & 0.69 & 0.73 & 0.71 \\
		Upper Wall & 0.40 & 0.69 & 0.49 & 0.57 \\
		Intestine & 0.20 & 0.64 & 0.22 & 0.33 \\
		Artery & 0.00 & 0.01 & 0.00 & 0.00 \\
		Bile & 0.00 & 0.00 & 0.00 & 0.00 \\
		Blood & 0.00 & 0.00 & 0.00 & 0.00 \\
		Black & 0.92 & 0.94 & 0.97 & 0.96 \\
		\textbf{Mean} & 0.26 & 0.40 & 0.33 & 0.33 \\  \hline
	\end{tabular}
	\caption{Results of our network on the m2caiSeg with all the 19 classes.}
	\label{tab:all_categories_results}
\end{table}

Again, we see similar trends as in the Single Instrument class case. The majority classes show good performance, especially Hook, Liver, Gallbladder, and Fat. Some categories, however, are completely inaccurately predicted, such as Unknown, Bipolar, Scissors, Irrigator, Trocars, Clip, Artery, Bile, and Blood. We can attribute the failure to the very few and often very difficult to discern instances of these classes in the training dataset.

Overall, we provide a baseline result in the problem domain. Our code and dataset are open-sourced to enable other researchers to contribute to the problem. We used the PyTorch Deep Learning framework \cite{paszke2017automatic} in our work, and our code is publicly available at \url{https://github.com/salmanmaq/segmentationNetworks}. Additionally, the annotated dataset can be can be accessed at \url{https://www.kaggle.com/salmanmaq/m2caiseg}

\subsection{EndoVis Robotic Scene Segmentation Challenge 2018}
The EndoVis Robotic Scene Segmentation challenge was hosted at MICCAI 2018. One of the main objectives of the challenge was to segment the different organs and instruments part of a laparoscopic image, which is quite similar to our work. The category labels and the surgical procedure covered are however different. We finetuned our network pre-trained on m2caiSeg on randomly sampled 80 \% of the EndoVis 2018 Robotic Scene Segmentation dataset training set for 42 epochs with a batch size of 4, and an initial learning of 0.005. The rest of the network hyperparameters were kept the same as our previous experiment. The datsaet is no longer publicly available, and since the test set labels were never made public, we evaluated our finetuned network on 20 \% of the training set used as a held out evaluation set. The dataset category labels and the respective performance scores of our network are given in Table \ref{tab:endovis2018_results}. 

\begin{table}[!htb]
	\centering
	\begin{tabular}{abbbb}
		\hline
		\rowcolor{Gray}
		\textsc{Class} & \textsc{IoU} & \textsc{Precision} & \textsc{Recall} & \textsc{F1 Score}\\
		\hline
		Background Tissue & 0.876 & 0.906 & 0.963 & 0.934 \\
		Instrument Shaft & 0.738 & 0.975 & 0.752 & 0.849 \\
		Instrument Clasper & 0.611 & 0.775 & 0.743 & 0.758 \\
		Instrument Wrist & 0.658 & 0.787 & 0.802 & 0.794 \\
		Kidney Parenchyma & 0.000 & 0.000 & 0.000 & 0.000 \\
		Covered Kidney & 0.758 & 0.886 & 0.840 & 0.863 \\
		Thread & 0.000 & 0.000 & 0.000 & 0.000 \\
		Clamps & 0.000 & 0.000 & 0.000 & 0.000 \\
		Suturing Needle & 0.000 & 0.000 & 0.000 & 0.000 \\
		Suction Instrument & 0.000 & 0.000 & 0.000 & 0.000 \\
		Small Intestine & 0.000 & 0.000 & 0.000 & 0.000 \\
		Ultrasound Probe & 0.000 & 0.000 & 0.000 & 0.000 \\ \hline
	\end{tabular}
	\caption{Results of our network on the EndoVis 2018 Robotic Scene Segmentation dataset.}
	\label{tab:endovis2018_results}
\end{table}

As can be observed in the Table \ref{tab:endovis2018_results}, our network performs well on the majority dataset classes, while it's performance suffers on the smaller objects such as needle and thread. This points out that our network is unable to pick up smaller structures in the scene, and that more advanced CNN architectures might be required. Additionally, it is important to note that the validation set contains images from the same sequences as the training set, which biases the results. However, the fact our network is able to generalize to the new dataset in fairly few epochs and to pick out the majority classes relatively easily highlights the strength of the proposed self-supervised pre-training and further finetuning scheme.

\section{Conclusion}
\label{sec:conclusion}
We introduced the problem of Laparoscopic Robotic Scene Segmentation in this paper, introduced the m2caiSeg dataset, and proposed a baseline network to tackle the problem. We also showed that the unsupervised pre-training for Semantic Segmentation is beneficial for Semantic Segmentation as image reconstruction enables learning features similar to the features learnt for Semantic Segmentation. Robotic scene segmentation is the first step towards autonomous surgical procedures as also depicted by the Endovis 2018 Robotic Scene Segmentation challenge, and thus the proposed work can play an important role towards this direction.

In the future, we plan to increase the instances of the minority classes, like Clip, Blood, Bile, and Artery; which are also critical to detect, in our dataset to achieve a better class balance. This would enable learning discriminating features for those classes as well and thus play a vital role in automating safety-critical surgical procedures.

UNet \cite{ronneberger2015unet} predictions retain the lower level information, such as edges and shapes. This shows that information sharing across the Encoder and Decoder networks can be useful. And that such an approach can also benefit from unsupervised pre-training. Secondly, powerful segmentation networks like DenseNet \cite{jegou2017one} can be trained if we have a larger dataset. Additionally, Dilated Convolutions \cite{yu2015multi} have been quite successful for Semantic Segmentation as they allow to capture context at multiple scales, thus providing a more global view of the input.

Recently, Pelt and Sethian \cite{pelt2017mixed} proposed a simple network that combines Dilated Convolutions with Dense blocks as in DenseNet. Here, they leverage the complementary properties of information sharing from lower layers to the later layers and information aggregation at multiple scales. We plan to explore this kind of approach for our case as well in the future.

Considering information shared between successive video frames can be also be useful to model temporal dependencies, where 3D CNNs \cite{ji20133d} and Recurrent Neural Networks can be helpful. Similarly, post-processing of Semantic Segmentation predictions due to spatial dependencies among neighboring pixels can be helpful to increase the network accuracy; but that usually comes at the cost of additional computational complexity and more complex network architectures.

\hspace{5pt}\\
\textbf{Funding:} This study was not funded by any institution or organization.

\textbf{Conflict of interest:} The authors declare that they have no conflict of interest.\\
\textbf{Ethical approval:} This article does not contain any studies with human participants or animals performed by any of the authors.

\end{document}